\pdfoutput=1

\documentclass[11pt]{article}

\usepackage[final]{acl}

\usepackage{times}
\usepackage{latexsym}
\usepackage{amsmath}
\usepackage{makecell}
\usepackage{multirow}
\usepackage{booktabs}
\usepackage{cleveref}
\usepackage{tcolorbox}
\usepackage[T1]{fontenc}
\usepackage[table]{xcolor}  
\definecolor{myblue}{RGB}{230, 242, 255}  
\definecolor{tableheadcolor}{gray}{0.92} 
\definecolor{tablerowcolor}{gray}{0.96} 

\usepackage[utf8]{inputenc}
\usepackage{CJKutf8}
\usepackage{microtype}
\usepackage{graphicx}

\usepackage{inconsolata}
\usepackage{graphicx}
\usepackage{subcaption}

\newcommand{\charYin}{{\CJKfamily{bsmi}蟫}}
\title{Who Wrote This Line? Evaluating the Detection of LLM-Generated Classical Chinese Poetry}

\author{
Jiang Li$^{1,2}$, 
Tian Lan$^{1,2}$, 
Shanshan Wang$^{3}$, 
Dongxing Zhang$^{1,2}$, 
Dianqing Lin$^{1,2}$,\\  
\textbf{Guanglai Gao}$^{1,2}$,
\textbf{Derek F. Wong}$^{3}$, 
\textbf{Xiangdong Su}$^{1,2}$\thanks{\ \ Corresponding Author}
 \\
$^1$ College of Computer Science, Inner Mongolia University, China \\ 
$^2$ National \& Local Joint Engineering Research Center of Intelligent Information\\ Processing Technology for Mongolian, China\\ 
$^3$ NLP²CT Lab, Department of Computer and Information Science, University of Macau, China \\
\texttt{velikayascarlet@gmail.com, cssxd@imu.edu.cn}}

\usepackage{fontawesome5}

\begin{document}
\maketitle
\begin{abstract}

The rapid development of large language models (LLMs) has extended text generation tasks into the literary domain. However, AI-generated literary creations has raised increasingly prominent issues of creative authenticity and ethics in literary world, making the detection of LLM-generated literary texts essential and urgent. While previous works have made significant progress in detecting AI-generated text, it has yet to address classical Chinese poetry. Due to the unique linguistic features of classical Chinese poetry, such as strict metrical regularity, a shared system of poetic imagery, and flexible syntax, distinguishing whether a poem is authored by AI presents a substantial challenge.  To address these issues, we introduce ChangAn, a benchmark for detecting LLM-generated classical Chinese poetry that containing total 30,664 poems, 10,276 are human-written poems and 20,388 poems are generated by four popular LLMs. Based on ChangAn, we conducted a systematic evaluation of 12 AI detectors, investigating their performance variations across different text granularities and generation strategies. Our findings highlight the limitations of current Chinese text detectors, which fail to serve as reliable tools for detecting LLM-generated classical Chinese poetry. These results validate the effectiveness and necessity of our proposed ChangAn benchmark. Our dataset and code are available at \url{https://github.com/VelikayaScarlet/ChangAn}.


\end{abstract}

\section{Introduction}
\begin{figure*}[t]  
    \centering
    \includegraphics[width=0.95\textwidth]{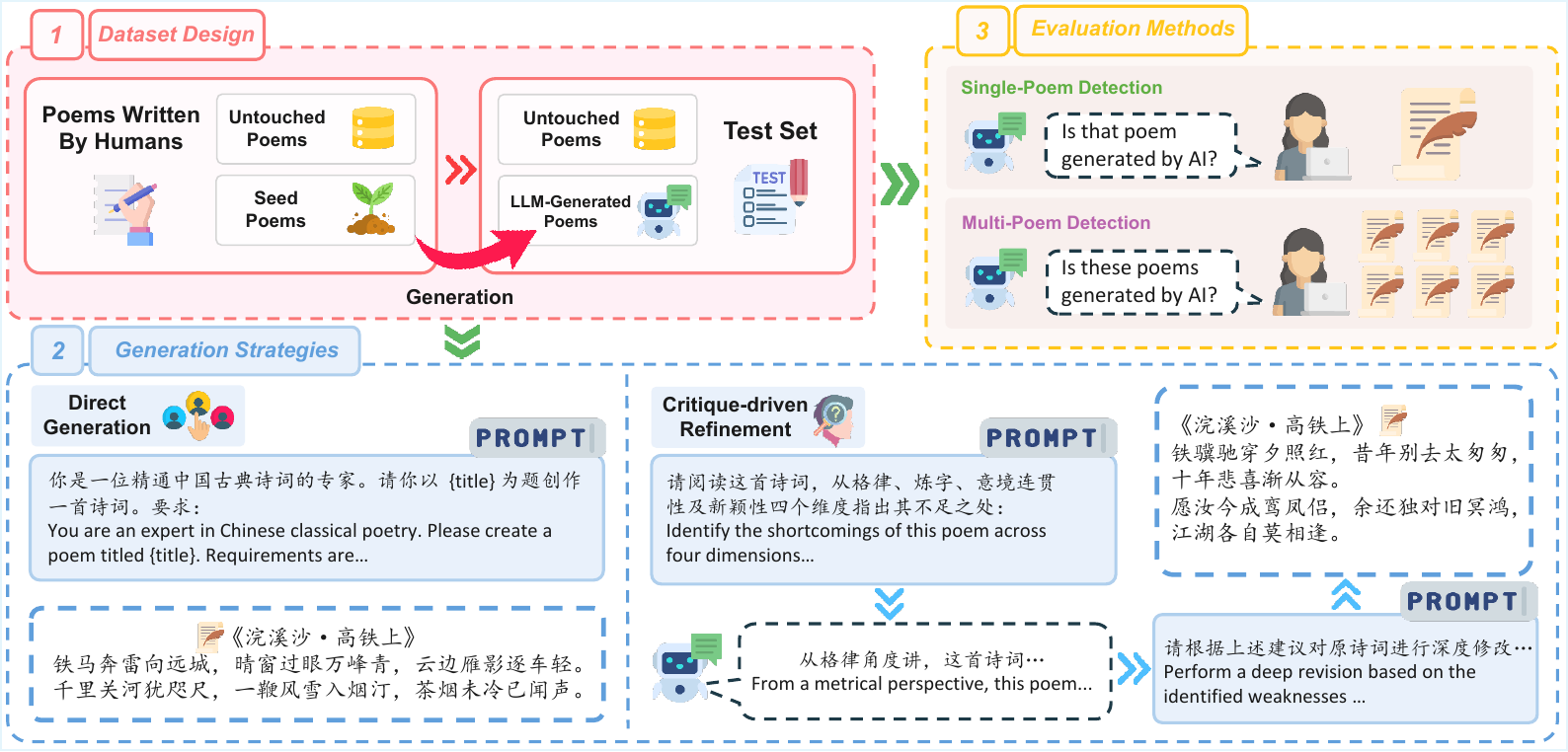} 
    \caption{The overall structure of our proposed benchmark}
    \label{fig:overall} 
\end{figure*}
Large language models~(LLMs) have demonstrated remarkable performance across a wide range~\citep{li-etal-2025-mutual,FDWR,HTC,mao2025robust,liu2025learning}. However, this powerful generative capability also entails potential risks: models may fabricate large amounts of false or even misleading content.
As a result, the automated detection of text generated by LLMs has become an increasingly urgent research need in the NLP field.

Chinese NLP has attracted much attention~\citep{xu2022guofeng,lan2025mcbe}. However, existing research on the detection of content generated by LLMs has primarily focused on modern languages and contemporary literary genres. Some highly literary forms such as English poetry~\citep{chen2024evaluating}, Haiku~\citep{jobanputra2025llm}, and Chinese classical~\citep{chen2025benchmarking} and modern poetry~\citep{wang-etal-2024-best,wang2026can}, which are increasingly becoming important extensions of LLMs' text generation capabilities, have remained largely underexplored in detection research.

\begin{CJK}{UTF8}{gbsn}
Currently, AI-generated content has given rise to increasingly prominent real-world issues in many realms~\citep{shi2024between}, and so has classical literature. Recent studies~\citep{CZXI202510010,wang-etal-2025-benchmarking} and media reports\footnote{AI诗歌成为获奖“钉子户”，“反AI诗歌联盟群”群主希望相关部门对AI作品投稿尽快出台相关政策} \footnote{《诗刊》副主编对AI诗歌投稿发出警告 “谁在写”引发文学圈深度探讨｜封面头条} \footnote{用AI诗作投稿：不是走捷径，而是绕远路} indicate that multiple poetry journals and literary platforms have publicly condemned the submission of AI-generated works without proper disclosure, reflecting growing concerns within the literary community regarding creative authenticity and authorship ethics. Reliance on human judgment alone has become insufficient to cope with the large-scale and low-cost generation of AI-generated content in this area.

In the Chinese literary tradition, classical poetry has long occupied a important position. Unlike modern poetry, which is relatively free in form, classical poetry is a highly formalized and strongly constrained literary genre, characterized by strict conventions governing line count, line length, tonal alternation, rhyme schemes, and parallel structures~\citep{downer1963tone,liu1988classical,xia2023necessity}, through which it achieves a distinctive sense of musicality and structural aesthetics.

However, these distinctive linguistic features create major challenges for general-purpose text detection methods when applied to classical Chinese poetry. Classical poetry displays strong regularities such as tonal alternation, rhyme patterns, and parallel structures~\citep{chen1979metrical,liu2015evolution}. 
As a result, detection models struggle to judge if these regularities reflect human adherence to poetic conventions or LLMs' imitation of learned patterns. 
Classical poetry also uses a widely shared set of poetic imagery. Many imagery terms appear often with both humans and AI models, causing overlap in lexical distributions~\citep{miao1991reading,hui2000comparative}. In addition, classical poetry has flexible word-class usage and frequent syntactic inversion~\citep{yu1971syntax,TSSF201703006}. For example, in "春风又绿江南岸", "绿 (green)" acts as a verb meaning make sth. green. In "竹喧归浣女", word order is inverted to meet metrical constraints. Such departures from modern Chinese grammar further complicate the task of capturing true linguistic patterns. Therefore,current AI-generated text detection benchmarks are inadequate for evaluating detector performance in this domain.

\end{CJK}
To address this gap, we construct and introduce ChangAn, a \underline{\textbf{Ch}}inese cl\underline{\textbf{a}}ssical poetry detection benchmark writte\underline{\textbf{n}} by lar\underline{\textbf{g}}e language models and modern hum\underline{\textbf{An}}s. ChangAn is designed to systematically evaluate the effectiveness of current LLMs and detection methods in distinguishing human-written and AI-generated classical poetry. ChangAn consists a total of 30,664 poems, including 10,276 high-quality poems written by 282 modern poets and poetry enthusiasts, as well as 20,388 poems generated by four mainstream LLMs, with a balanced distribution across models, ensuring broad thematic coverage and stylistic diversity. 
Additionally, we evaluate four representative LLMs and eight detection baselines across distinct scenarios based on text granularity and generation strategies. Notably, we also investigate the dual role of LLMs as both content generators and zero-shot detectors to explore their capacity for self-detection in the context of classical poetry. The overall structure of ChangAn can be found in Figure~\ref{fig:overall}. Our key contributions are as follows:


\begin{itemize}
    \item \textbf{Comprehensive Dataset} We introduce \textbf{ChangAn}, the first specialized benchmark for detecting LLM-generated classical Chinese poetry, featuring a diverse collection of contemporary human-written and LLM-generated poems.
    \item \textbf{Evaluation Benchmark} We establish a novel benchmark that incorporates diverse text granularities and generation strategies, enabling a holistic assessment of detection methods across various creative scenarios.
    \item \textbf{Experimental Analysis} We conduct a comprehensive evaluation of different detectors, exposing the limitations of them and offering new insights into AI detection for classical literary languages.
\end{itemize}

\section{Related Works}

\subsection{Classical Chinese poetry Generation}

Classical Chinese poetry is characterized by rigorous metrical patterns and condensed imagery, features that distinguish its generation from standard natural language tasks. Early rule-based systems~\citep{yan2013poet} failed to balance strict formal constraints like tonal balance and rhyming with semantic naturalness and artistic expressiveness.

The advent of deep learning led to the widespread adoption of Recurrent Neural Networks (RNNs) in poetry generation~\citep{wang2016chinese,yi2017generating}, which significantly enhanced semantic coherence and thematic consistency. In the Transformer era, LLMs have further bolstered generative capabilities by leveraging large-scale corpora to internalize complex linguistic structures~\citep{luo2021chinese,huang2025poembert}. Furthermore, recent studies have achieved substantial progress in addressing long-standing challenges, including emotional and stylistic control~\citep{shao2021sentiment} as well as intricate formatting and metrical constraints~\citep{yu2024charpoet,song2025mixsong}.


\subsection{LLM-Generated and Human-Written Poetry}

As generative artificial intelligence permeates literary creation, discussions regarding the boundaries between LLMs and human authorship have intensified. In studies focusing on English poetry, \citet{kobis2021artificial} examined the stylistic quality and subjective credibility of texts generated by GPT-2. Their work introduced two pivotal concepts: Humans-out-of-the-loop (HOTL) and Humans-in-the-loop (HITL). Experimental results indicated that human participants face significant challenges in identifying HITL poetry involving manual intervention. This conclusion was further supported by \citet{gunser2022pure} and \citet{porter2024ai}, whose research revealed substantial cognitive limitations in laypeople when distinguishing between AI-generated verses and works by renowned poets.

In the domain of short-form poetry, \citep{hitsuwari2023does} conducted comparative experiments on Japanese Haiku. Given the extremely concise "5-7-5" syllable structure and the inherent imagery ambiguity of Haiku~\citep{hitsuwari2023does,tateishi2025we}, participants could not effectively differentiate between human-made and AI-generated works. Notably, high-quality AI Haikus were frequently misidentified as human-written, which underscores the high fidelity of AI in constrained text generation. Furthermore, \citet{wang-etal-2025-benchmarking} established the first benchmark for detecting modern Chinese poetry. This study highlighted the impact of AI-generated content on the literary ecosystem and suggested that intrinsic linguistic traits, particularly style, remain the most difficult dimensions for detection, whereas poems with explicit emotional expressions are relatively easier to identify.

The detection of classical Chinese poetry presents inherent difficulties due to short text length, fixed linguistic patterns, and sparse structural information~\citep{chen2025benchmarking}. First, standard formats  restrict each verse to very few tokens, thereby limiting the discriminative features available for detection methods. Second, the widespread use of shared imagery, classical allusions, and specific rhetorical devices across the genre makes more difficult to distinguish the creation between human and AI. Consequently, this has become one of the most challenging detection tasks within the NLP field.

\section{The Dataset}
\begin{table*}[ht]
\centering
\rowcolors{2}{white}{tablerowcolor} 

\begin{tabular}{lccccc}
\toprule
\rowcolor{tableheadcolor} 
\textbf{Author} & \textbf{Num. Poems} & \textbf{Num. Sents} & \textbf{Avg. S/P} & \textbf{Avg. C/S} & \textbf{Total Characters} \\ 
\midrule
Human           & 10,276 & 105,828 & 10.30 & 5.70 & 603,034 \\
\hline
Doubao Seed-1.6 & 5,136  & 48,232  & 9.39  & 6.13 & 295,742 \\
DeepSeek-V3.2     & 5,138  & 51,858  & 10.09 & 5.89 & 305,285 \\
Kimi-K2         & 5,017  & 51,129  & 10.19 & 5.92 & 302,812 \\
GPT-4.1         & 5,097  & 50,713  & 9.95  & 5.83 & 295,503 \\ 
Overall & 30,664  & 307,760  & 10.04  & 5.86 & 1,802,646 \\ 
\bottomrule
\end{tabular}
\small 
\caption{Statistical distribution of the ChangAn dataset. Abbreviations include Num. Poems (Number of Poems), Num. Sents (Number of Sentences), Avg. S/P (Average Sentences Per Poem), and Avg. C/S (Average Characters Per Sentence).}
\label{tab:dataset-stats}
\end{table*}
\subsection{Dataset Statistics}
High-quality data constitutes the foundation of any evaluation benchmark~\citep{wu2025survey,fang2025lastingbench,peng2025swe}. However, there is currently a lack of publicly available datasets featuring classical poetry authored by modern poets. To fill this gap, we constructed the ChangAn dataset, a novel corpus comprising 10,276 high-quality classical poems by 282 modern authors and 20,388 samples generated by four mainstream LLMs. To the best of our knowledge, this is the first classical poetry dataset to simultaneously incorporate both modern human compositions and LLM-generated works. It includes 17193 Ci, 3773 jueju and  9699 lvshi. Detailed statistical information for the ChangAn dataset can be found in Table~\ref{tab:dataset-stats}. We provide poem examples in \textbf{Appendix~\ref{app:examples-of-poetry}.}

\subsection{Poetry Written by Humans}
We curated a dataset of 10,276 high-quality classical poems authored by 282 modern poets and poetry enthusiasts. Specifically, the data were collected from two primary sources. First, we crawled poems from popular online social platforms and communities, such as Xiaohongshu and Baidu Tieba. Second, we selected high-quality works from professional literary publications and themed anthologies. These sources represent a wide range of contemporary poetry creation, where modern poets actively publish their works and exchange creative insights. These works strictly adhere to traditional norms regarding metrical patterns, imagery, classical allusions, and linguistic style, while simultaneously incorporating innovative elements such as modern imagery. Consequently, this collection provides an authentic reflection of contemporary human proficiency in classical composition. Importantly, all authors' names (or pseudonyms), are preserved within the dataset to acknowledge intellectual property and ensure complete transparency.

Our decision to utilize classical poetry by modern authors rather than ancient literary figures and poets stems from the fact that most ancient texts have already been incorporated into the pre-training corpora of LLMs. If ancient works were used to evaluate AI detection methods, models could easily determine the source by "memorizing" specific text segments, thereby failing to accurately reflect their true discriminative capabilities. In contrast, modern classical poetry effectively mitigates the bias caused by data leakage. This approach ensures that the detection task remains focused on its core objective: whether a model can distinguish between human-written and AI-generated texts based on stylistic, structural, and linguistic features rather than mere memorization. 

\subsection{Poetry Generated by LLMs}
In addition to human-written classical poetry, our dataset incorporates 20,388 samples generated by four prominent LLMs: DeepSeek-V3.2~\citep{liu2024deepseek}, GPT-4.1~\citep{achiam2023gpt}, Kimi-K2~\citep{team2025kimi}, and Doubao Seed-1.6~\citep{team2025seed1}.
Furthermore, we designed two distinct prompting strategies to simulate direct generation based on a given topic and critique-driven refinement. During the data generation phase, we randomly selected 2,569 human-written poems from the original corpus as a seed set (1/4 of all poems). These samples guided the LLMs in generation based on their specific titles, styles, and genres. The remaining 7,707 poems were excluded from the generation process and instead mixed with the AI-generated works to construct a comprehensive test set comprising 28,096 poems. This design provides a robust baseline for subsequent discrimination experiments by incorporating pristine human samples that the models never encountered during the generation phase. The prompts of direct generation $P_g$ and critique-driven refinement $P_c$ can be found in \textbf{Appendix~\ref{app:prompts}}.

For the generation parameters, we adopted the recommended temperatures for creative writing: 1.5 for DeepSeek-V3.2 and 1.0 for the other models. To better analyze the stylistic differences between human-written and AI-generated poetry, we provide word cloud diagrams and high-frequency image statistics for both AI and human subsets in \textbf{Appendix \ref{app:imagerywords}}.

\section{Evaluation Methods} Accurately evaluating the discriminative capability of detection methods for LLM-generated poetry presents a significant challenge. Relying solely on the direct detection of fully AI-generated works fails to capture the actual efficacy of these tools in complex, real-world scenarios. This limitation arises from the diverse ways LLMs are utilized: they can either create entire poems independently or be employed to polish and revise human-written drafts. Consequently, we systematically evaluate existing detection methods across two dimensions: text granularity and generation strategy.
\begin{CJK}{UTF8}{gbsn}
\subsection{Text Granularity} 
To evaluate performance boundaries across different data scales, we categorize the detection tasks into \textbf{Single-Poem Detection (SPD)} and \textbf{Multi-Poem Detection (MPD)}.

We select 6 and 12 poems for the MPD setting to reflect both cultural norms. A collection of 6 poems is a common and cohesive unit in classical Chinese poetry such as 《三吏三别》(Three Officials and Three Partings), 《塞下曲六首》(Six Songs of Under the Frontier), 《戏为六绝句》(Six Quatrains Written Playfully), and 《上元夜六首》(Six Poems on the Lantern Festival Night), making it a natural choice for evaluation. To report performance under a larger scale, we additionally adopt 12 poems (twice the size of 6), allowing us to analyze the impact of doubling the input data on detection efficacy.

\textbf{Single-Poem Detection} This setting focuses on the most common form of poetry sharing. Because Chinese classical poems are short, they provide limited semantic and stylistic features for detectors. This reflects how detection methods perform in real-world contexts such as social media or submission platforms.

\textbf{Multi-Poem Detection} This setting simulates the publication of poetry collections, in which an author’s lexical choices and imagery tend to show stylistic similarity across works. We therefore evaluate ensembles of 6 and 12 poems to assess how sample size affects detection accuracy.

\begin{table*}[t]
  \centering
  \resizebox{0.95\textwidth}{!}{ 
  \begin{tabular}{lcccccc}
    \hline
    \rowcolor{blue!15} \textbf{Detectors} & \textbf{Acc.} & \textbf{Recall (AI)} & \textbf{Recall (Human)} & \textbf{Macro-Recall} & \textbf{Macro-F1} \\
    \hline
    
    Deepseek-V3.2   & 39.56 & 31.79 & 60.96 & 46.37 & 39.42 \\
    \rowcolor{blue!5} Kimi-K2 & 31.98 & 16.06 & 75.83 & 45.94 & 31.30 \\
    Doubao Seed-1.6 & 33.80 & 22.08 & 66.07 & 44.07 & 33.96 \\
    \rowcolor{blue!5} GPT-4.1 & 26.02 & 5.10  & 83.63 & 44.37 & 23.31 \\
    LLM-Detector-Small-zh & 29.38 & 3.86  & 98.74 & 51.30 & 25.16 \\
    \hline
    \rowcolor{gray!10} \textbf{Average} & 32.15 & 15.78 & 77.05 & 46.41 & 30.63 \\
    \hline
  \end{tabular}}
  
  \caption{Overall performance comparison of LLM detector. The metrics include Accuracy (Acc.), Recall for AI-generated and Human-written poetry, and Macro-F1.}
  \label{tab:overall-performance-of-dbd}
\end{table*}

\begin{table}[t]
  \centering
  \small
  \begin{tabular}{lcc}
    \hline
    \rowcolor{blue!15} \textbf{Detectors} & \textbf{AUROC} & \textbf{Macro-F1} \\
    \hline
    Log-Likelihood & 83.94 & 73.77 \\
    \rowcolor{blue!5}Log-Rank & 85.86 & 75.56 \\
    LRR & 71.09 & 64.81 \\
    \rowcolor{blue!5}Roberta & 95.03 & 86.18 \\
    Zh-v3 & 72.37 & 65.44 \\
    \rowcolor{blue!5}Zh-v3-short & 72.25 & 65.06 \\
    Fast-DetectGPT & 49.67 & 49.35 \\
    \hline
    \rowcolor{gray!10} \textbf{Average} & 75.89 & 67.14 \\
    \hline
  \end{tabular}
  \caption{Performance comparison of detection methods. Note that "Zh-v3" and "Zh-v3-short" refer to the AIGC-Detector-v3 (Zh) and its short-text variant, respectively.}
  \label{tab:overall-comparison-of-pbd}
\end{table}
\subsection{Generation Strategy} 
We further distinguish between two strategies to simulate the varying depths of LLM involvement in the creative process.

\textbf{Direct Generation} This strategy refers to a setting in which text is generated directly by an LLM without any post-processing. It serves to evaluate whether detection methods can identify patterns that are purely generated by LLMs.

\textbf{Critique-driven Refinement} This strategy simulates the logical process of "推敲 (deliberating and crafting)" in human composition. It focuses on how LLMs optimize text through self-evaluation and targeted adjustments. This mode poses a greater challenge to the discriminative capabilities of detectors compared to direct generation. We have shown how it works in \textbf{Appendix~\ref{app:models-critique}}.
\end{CJK}

\section{Experimental Setup}

\textbf{Decision-based Detectors} We selected four representative LLMs for our experiments: DeepSeek-V3.2, GPT-4.1, Kimi-K2, and Doubao Seed-1.6. Through grouped experiments, we aim to analyze whether these models can identify their own generated poems, thereby verifying their self-recognition capabilities. The prompts we used in this task can be found in \textbf{Appendix \ref{app:prompts}}.

\textbf{Probability-based Detectors} We evaluated a variety of methods on the ChangAn dataset following standard configurations from prior research. For Log-Likelihood~\citep{solaiman2019release}, Log-Rank~\citep{gehrmann2019gltr}, and LRR~\citep{su2023detectllm}, we utilized Qwen2.5-3B~\citep{qwen2025qwen25technicalreport} as the base scoring model. For Fast-DetectGPT~\citep{bao2023fast}, we employed Qwen2.5-3B as the reference model and Qwen2.5-7B as the scoring model. Regarding supervised detectors, we used AIGC-Detector-V3~\citep{tian2023multiscale} (including both the Zh-v3 version and its short-text optimized Zh-v3-short variant) and LLM-Detector-Small-zh~\citep{wang2024llm} model. All baselines were configured with their default parameters. Additionally, Roberta~\citep{liu2019roberta} is also introduced, which has prominent performance in classification tasks. We fine-tuned the Chinese Roberta model with 3 epochs, a learning rate of 1e-4, and a batch size of 16. When training the RoBERTa model, we randomly split the entire dataset into training set, validation set, and test set at a ratio of 8:1:1.

To ensure reliability, all experiments were conducted three times on 2 A100 GPUs, each with 80G memory, and we reported the average values as our final results.

\textbf{Metrics} For decision-based detectors, where logit values are inaccessible, we report Accuracy, Recall for AI, Recall for Human, and Macro-F1 based on their discrete classification outputs. For probability-based detectors, we follow previous work and report AUROC and Macro-F1 scores.

\section{Results and Analysis}

\begin{table*}[t]
  \centering
  
  \resizebox{\textwidth}{!}{
  \begin{tabular}{lcccccccccc}
    \hline
    \rowcolor{blue!15} 
    & \multicolumn{2}{c}{\textbf{Deepseek-V3.2}} & \multicolumn{2}{c}{\textbf{Kimi-K2}} & \multicolumn{2}{c}{\textbf{Doubao Seed-1.6}} & \multicolumn{2}{c}{\textbf{GPT-4.1}} & \multicolumn{2}{c}{\textbf{Overall}} \\
    
    \rowcolor{blue!15} \multirow{-2}{*}{\textbf{Detectors}} & AUROC & Macro-F1 & AUROC & Macro-F1 & AUROC & Macro-F1 & AUROC & Macro-F1 & AUROC & Macro-F1 \\
    \hline
    
    Log-Likelihood & 82.39 & 75.30 & 84.62 & 77.42 & 84.62 & 77.22 & 84.19 & 76.85 & 83.94 & 73.77 \\
    \rowcolor{blue!5}Log-Rank & 83.55 & 76.86 & 86.78 & 80.06 & 86.60 & 79.16 & 86.60 & 79.53 & 85.86 & 75.56 \\
    LRR & 72.27 & 66.56 & 70.98 & 65.98 & 70.89 & 65.38 & 70.15 & 64.89 & 71.09 & 64.81 \\
    \rowcolor{blue!5}Roberta & 93.61 & 81.69 & 93.90 & 81.89 & 97.95 & 84.12 & 94.64 & 82.52 & 95.03 & 86.18 \\
    
    Zh-v3 & 77.16 & 69.91 & 66.11 & 61.49 & 69.42 & 63.74 & 79.38 & 71.48 & 72.37 & 65.44 \\
    \rowcolor{blue!5}Zh-v3-short & 77.62 & 70.15 & 65.60 & 60.56 & 72.82 & 66.68 & 77.74 & 69.92 & 72.25 & 65.06 \\
    Fast-DetectGPT & 53.60 & 46.69 & 50.92 & 44.99 & 52.10 & 45.92 & 41.70 & 40.71 & 49.67 & 49.35 \\
    
    \hline
    \rowcolor{gray!10} \textbf{Average} & 77.17 & 69.59 & 74.13 & 67.48 & 76.34 & 68.89 & 76.34 & 69.41 & 75.63 & 68.60 \\
    \hline
  \end{tabular}}
  \caption{AUROC and Macro-F1 comparison across LLM generators.  Note that "Zh-v3" and "Zh-v3-short" refer to the AIGC-Detector-v3 (Zh) and its short-text variant, respectively.}
  \label{tab:llm-detailed}
\end{table*}
\subsection{Overall Performance}

Table~\ref{tab:overall-performance-of-dbd} illustrates the performance of various decision-based detectors. The results indicate that most detectors perform poorly and tend to misclassify both AI-generated and human-written poems as human creations. DeepSeek-V3.2 achieves the highest performance among the tested detectors. In contrast, GPT-4.1 shows the weakest performance. It is noteworthy that all Chinese-developed models outperform GPT-4.1, which likely stems from the differences in their respective training corpora and exposure to classical Chinese literature.

Unexpectedly, LLM-Detector-Small-zh, which is specifically optimized for the Chinese language, underperforms most detectors. This suggests that existing decision-based detectors trained on general text possess extremely limited generalization capabilities when applied to specialized genres such as classical poetry.

Table~\ref{tab:overall-comparison-of-pbd} presents the results of probability-based detectors. Several key observations can be made. First, statistical methods demonstrate a significant advantage in poetry detection compared to the decision-based detectors discussed previously. Log-Likelihood and Log-Rank achieve robust performance with AUROC scores exceeding 80\%. This suggests that while AI-generated poetry may differ semantically from human-written poems, it exhibits distinguishable patterns in word and imagery usage, as illustrated by our analysis in Appendix~\ref{app:imagerywords}. Notably, the fine-tuned Roberta model achieves the highest AUROC of 95.03\%, proving the potential of supervised learning in classical poetry detection.

Second, the Fast-DetectGPT fails completely. Its AUROC of 49.67\% is nearly equivalent to random guessing. This indicates that the negative curvature hypothesis relied upon by Fast-DetectGPT does not hold under the highly constrained linguistic structures of classical poetry. Because the creative space of classical poetry is constrained by strict rhythm and parallelism, standard perturbation analysis may struggle to distinguish statistical artifacts from the few viable token alternatives.

Finally, specialized tools like AIGC-Detector-V3 (Zh-v3 and Zh-v3-short) show limited efficacy, as their performance remains substantially lower than that of basic statistical methods.

\subsection{Can LLMs Identify Their Own Poetic Creations?}

\begin{figure}[t]
\centering
\scalebox{0.99}{
  \includegraphics[width=\columnwidth]{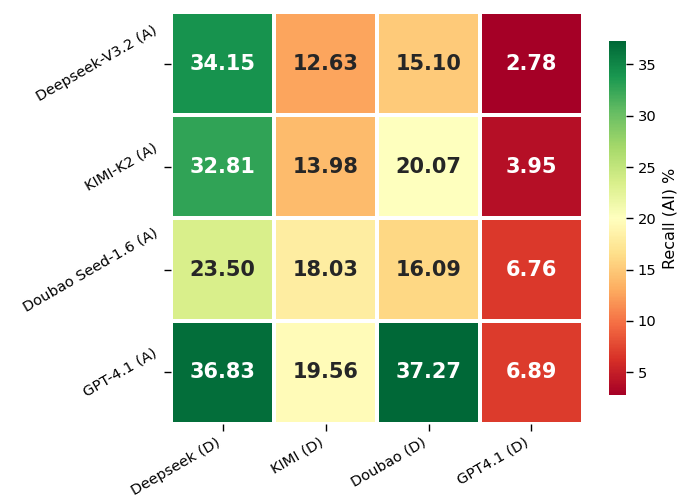}}
  \caption{Cross-model recall performance for poetry detection. (A) denotes Author models; (D) denotes Detector models. Numerical values of Recall (AI) represent the percentage of correctly identified AI samples.}
  \label{fig:self-detection}
\end{figure}

Figure~\ref{fig:self-detection} illustrates cross-model recall in poetry detection. Two key observations emerge regarding models' ability to identify their own outputs.

First, most models show little advantage in self-recognition, unlike in general text domains. \citet{panickssery2024llm} found that for common texts, a model's self-recognition correlates linearly with self-preference bias, reflecting reliance on an "identity signal." In classical poetry, this signal appears significantly weakened: Doubao Seed-1.6 achieves only 16.09\% recall on its own samples, much lower than its 37.27\% recall when detecting GPT-4.1, indicating that the strict constraints of poetic form effectively mask the statistical patterns typically exploited by models.

Second, the discriminative capability may depends more on domain-specific modeling depth than on model identity. DeepSeek-V3.2 exhibits the most robust and consistent detection performance across all author groups, while GPT-4.1 fails to recognize AI-generated poetry regardless of the source, with recall rates consistently below 7\%. Given that these models struggle to identify their own work, their evaluative judgments in this domain are unlikely to be driven by the self-recognition-based bias identified by \citet{panickssery2024llm}. Instead, the results indicate that AI-generated poetry has reached a level of stylistic maturity where its machine traces are indistinguishable even to the originating models.

\subsection{Which Model's Poems is the Easiest to Identify?}

Table~\ref{tab:llm-detailed} presents a detailed evaluation of detection performance across various generative models. Kimi-K2 proves to be the most "stealthy" model. It yields the lowest average AUROC (74.13\%) and Macro-F1 (67.48\%), suggesting a higher degree of integration with the natural distribution of human-written poetry. However, the results demonstrate that the average AUROC scores across the four sources of generated poetry exhibit only modest variations. This level of comparability suggests that different LLMs tend to leave a similar density of statistical artifacts when constrained by the rigid structures of classical Chinese poetry. 

The consistently high performance of the fine-tuned Roberta model across all groups, particularly its 97.95\% AUROC on Doubao Seed-1.6, demonstrates that supervised detectors can effectively capture shared machine traits even when the source models differ in architecture and scale.

\subsection{Impact of Generation Strategies}
Table~\ref{tab:generation-vs-critique} and Table~\ref{tab:strategy-comparison-detailed} present the performance of various detectors across two distinct generation strategies: direct generation and critique-driven refinement. The results indicate that the refinement process generally enhances the detection difficulty of AI-generated poetry generally, although its impact varies slightly across different detectors.
\begin{table}[t]
  \centering
  \resizebox{\columnwidth}{!}{ 
    \begin{tabular}{lcc} 
      \hline
      \rowcolor{blue!15} 
      \textbf{Detectors} & \textbf{Recall (Gene.)} & \textbf{Recall (Crit.)} \\ 
      \hline
      
      Doubao Seed-1.6       & 21.88 & 22.29 (+0.41) \\
      \rowcolor{blue!5} 
      Deepseek-V3.2         & 36.72 & 26.87 ($-$9.85) \\
      Kimi-K2               & 18.33 & 13.79 ($-$4.54) \\
      \rowcolor{blue!5} 
      GPT-4.1               & 4.64  & 5.56 (+0.92) \\
      LLM-Detector & 52.13 & 50.48 ($-$1.65) \\
      \hline
      \rowcolor{gray!10} 
      \textbf{Average}      & 26.74 & 23.80 ($-$2.94) \\
      \hline
    \end{tabular}
  }
  \caption{Performance comparison of decision-based detectors across different generation strategies, we report the recall of AI-genereted poetry. LLM-Detector denotes the LLM-Detector-Small-zh. Values in parentheses denote the recall shift of the latter relative to the former.}
  \label{tab:generation-vs-critique}
\end{table}

\begin{table}[t]
  \centering
  \resizebox{\columnwidth}{!}{ 
  \begin{tabular}{lcccc}
    \hline
    \rowcolor{blue!15} 
    & \multicolumn{2}{c}{\textbf{Generation}} & \multicolumn{2}{c}{\textbf{Critique}} \\
    
    \rowcolor{blue!15} \multirow{-2.2}{*}{\textbf{Detectors}} & \textbf{AUROC} & \textbf{Macro-F1} & \textbf{AUROC} & \textbf{Macro-F1} \\
    \hline
    
    Log-Likelihood          & 93.33 & 86.54 & 74.70 ($-$18.63) & 68.61 ($-$17.93) \\
    \rowcolor{blue!5} 
    Log-Rank                & 94.73 & 88.53 & 77.12 ($-$17.61) & 70.65 ($-$17.88) \\
    LRR                     & 77.76 & 71.13 & 64.53 ($-$13.23) & 61.21 ($-$9.92) \\
    \rowcolor{blue!5} 
    Roberta                 & 97.69 & 87.44 & 92.38 ($-$5.31) & 83.58 ($-$3.86) \\
    Zh-v3                   & 75.62 & 69.18 & 70.49 ($-$5.13) & 64.88 ($-$4.30) \\
    \rowcolor{blue!5} 
    Zh-v3-short             & 76.36 & 69.59 & 70.61 ($-$5.75) & 65.01 ($-$4.58) \\
    Fast-DetectGPT          & 51.52 & 49.40 & 47.85 ($-$3.67) & 47.35 ($-$2.05) \\
    \hline
    \rowcolor{gray!10} 
    \textbf{Average}        & 80.93 & 74.54 & 71.10 ($-$9.83) & 65.90 ($-$8.64) \\
    \hline
  \end{tabular}}
  
  \caption{Performance comparison of probability-based detectors across generation strategies. We report the AUROC and Macro-F1 score. Parentheses indicate the decrease in performance from direct generation to critique-driven refinement.}
  \label{tab:strategy-comparison-detailed}
\end{table}

\begin{table*}[t]
  \centering

  \resizebox{\textwidth}{!}{
  \begin{tabular}{lcccccccccccc}
    \hline
    \rowcolor{blue!15} 
    & \multicolumn{4}{c}{\textbf{SPD}} & \multicolumn{4}{c}{\textbf{MPD-6}} & \multicolumn{4}{c}{\textbf{MPD-12}} \\
    
    \rowcolor{blue!15} \multirow{-2.5}{*}{\textbf{Detectors}} 
    & \textbf{Acc} & \textbf{R-AI} & \textbf{R-Hu} & \textbf{M-F1} 
    & \textbf{Acc} & \textbf{R-AI} & \textbf{R-Hu} & \textbf{M-F1} 
    & \textbf{Acc} & \textbf{R-AI} & \textbf{R-Hu} & \textbf{M-F1} \\
    \hline
    
    Deepseek-V3.2     & 39.56 & 31.79 & 60.96 & 39.42 & 61.75 & 74.51 & 26.58 & 50.54 & 62.83 & 74.18 & 31.53 & 52.83 \\
    \rowcolor{blue!5} 
    Kimi-K2           & 31.98 & 16.06 & 75.83 & 31.30 & 50.84 & 42.81 & 72.97 & 50.12 & 54.92 & 55.23 & 54.05 & 51.61 \\
    Doubao Seed-1.6   & 33.80 & 22.08 & 66.07 & 33.96 & 55.52 & 49.35 & 72.52 & 54.21 & 54.20 & 48.69 & 69.37 & 52.79 \\
    \rowcolor{blue!5} 
    GPT-4.1           & 26.02 & 5.10  & 83.63 & 23.31 & 69.54 & 92.48 & 6.31  & 45.80 & 71.94 & 94.12 & 10.81 & 50.07 \\
    LLM-Detector-S-zh & 29.38 & 3.86  & 98.74 & 25.16 & 42.86 & 31.57 & 84.24 & 42.60 & 28.73 & 15.67 & 85.51 & 28.65 \\
    \hline
    \rowcolor{gray!10} 
    \textbf{Average}  & \textbf{32.15} & \textbf{15.78} & \textbf{77.05} & \textbf{30.63} 
                      & \textbf{56.10} & \textbf{58.14} & \textbf{52.52} & \textbf{48.65} 
                      & \textbf{54.52} & \textbf{57.58} & \textbf{50.25} & \textbf{47.19} \\
    \hline
  \end{tabular}}
  
  \caption{Performance comparison across MPD, SPD-6, and SPD-12 strategies. R-AI and R-Hu denote Recall for AI and Human samples, respectively. M-F1 stands for Macro-F1 score.}
  \label{tab:m-s-pd-dbd}
\end{table*}

\begin{table*}[t]
  \centering
  \begin{tabular}{lcccccc}
    \hline
    \rowcolor{blue!15} 
    & \multicolumn{2}{c}{\textbf{SPD}} & \multicolumn{2}{c}{\textbf{MPD-6}} & \multicolumn{2}{c}{\textbf{MPD-12}} \\
    
    \rowcolor{blue!15} \multirow{-2.5}{*}{\textbf{Detectors}} 
    & \textbf{AUROC} & \textbf{M-F1} 
    & \textbf{AUROC} & \textbf{M-F1} 
    & \textbf{AUROC} & \textbf{M-F1} \\
    \hline
    
    Log-Likelihood          & 83.94 & 73.77 & 98.90 & 95.71 & 99.96 & 99.08 \\
    \rowcolor{blue!5} 
    Log-Rank                & 85.86 & 75.56 & 99.29 & 96.97 & 100.0 & 99.70 \\
    LRR                     & 71.09 & 64.81 & 99.63 & 97.24 & 100.0 & 99.70 \\
    \rowcolor{blue!5} 
    Roberta                 & 95.03 & 86.18 & 99.96 & 84.89 & 99.99 & 84.76 \\
    Zh-v3                   & 72.37 & 65.44 & 87.98 & 80.51 & 89.90 & 83.43 \\
    \rowcolor{blue!5} 
    Zh-v3-short             & 72.25 & 65.06 & 82.73 & 74.60 & 84.59 & 77.28 \\
    Fast-DetectGPT          & 49.67 & 49.35 & 45.16 & 44.27 & 45.29 & 42.22 \\
    \hline
    \rowcolor{gray!10} 
    \textbf{Average}        & \textbf{75.74} & \textbf{68.59} & \textbf{87.95} & \textbf{82.03} & \textbf{88.68} & \textbf{83.74} \\
    \hline
  \end{tabular}
  
  \caption{Comparison of detection performance (AUROC and Macro-F1) across different window strategies. M-F1 stands for Macro-F1 score. Zh-v3 refers to AIGC-Detector-v3 (Zh-v3).}
  \label{tab:m-s-pd-pbd}
\end{table*}
First, as the Table~\ref{tab:generation-vs-critique} shows, for decision-based detectors, critique-driven refinement has a modest but meaningful effect. On average, recall drops from 26.74\% to 23.80\%, though individual models exhibit varied responses: DeepSeek-V3.2 decreases by 9.85\%, while Doubao Seed-1.6 and GPT-4.1 increase slightly by 0.41\% and 0.92\%, respectively. This indicates that for certain detectors, the refinement process can even introduce patterns that are more detectable.

In contrast, the impact of the critique-driven refinement strategy is even more pronounced for probability-based detectors. According to Table~\ref{tab:strategy-comparison-detailed}, the average AUROC across these methods falls from 80.93\% to 71.10\%. Methods like Log-Likelihood and Log-Rank, which perform well on direct generation, experience substantial performance degradation of approximately 17\% to 18\% in AUROC when evaluating samples from critique-driven refinement. This provides evidence that the refinement process effectively reduces the identifiable statistical artifacts that characterize raw LLM outputs.

While the fine-tuned Roberta model remains the most robust detector, even its performance shows a slight decrease in AUROC from 97.69\% to 92.38\%. This clear adversarial advantage afforded by critique-driven refinement underscores a significant challenge for current detection methodologies.

\subsection{Influence of Text Granularity}
Table~\ref{tab:m-s-pd-dbd} and Table~\ref{tab:m-s-pd-pbd} illustrate the detection performance across different text granularities. The results demonstrate that Multi-Poem Detection (MPD) significantly outperforms Single-Poem Detection (SPD), as the aggregation of multiple poetic samples provides a richer set of features detectors.

GPT-4.1 presents a unique case of extreme classification shift. In the SPD setting, it achieves an R-Hu of 83.63\% but an R-AI of only 5.10\%, indicating a strong bias towards classifying its outputs as human-written. However, this bias reverses dramatically in MPD-6, where R-AI surges to 92.48\% while R-Hu drops to 6.31\%. This suggests that for GPT-4.1, the accumulation of text leads to a violent shift from human-leaning to AI-leaning predictions, a phenomenon not observed as intensely in other models.

We found that the performance gain from increasing text length exhibits clear diminishing returns. As shown in the average metrics in Table~\ref{tab:m-s-pd-dbd}, the transition from SPD to MPD-6 yields a substantial increase in average Accuracy (from 32.15\% to 56.10\%) and R-AI (from 15.78\% to 58.14\%). However, further increasing the granularity to MPD-12 results in negligible improvements, and in some cases, a slight decline in performance. Same but more evident trend can be observed in Table~\ref{tab:m-s-pd-pbd}. The average AUROC across these methods increases from 75.74\% in SPD to 87.95\% in MPD-6, and further to 88.68\% in MPD-12. Both Log-Rank and LRR achieve almost perfect AUROC scores. 

This confirms that the statistical signals of AI-generated poetry, while subtle in isolated samples, become highly distinguishable as the available text volume increases. However, this performance gain follows a pattern of diminishing returns. The results indicate that a  size of 6 poems already captures the majority of identifiable features, so more poems in the MPD-12 setting yields only marginal improvements.

\section{Conclusion}
In this paper, we propose ChangAn, the first specialized benchmark for classical Chinese poetry detection. Comprising 30,664 samples, ChangAn features diverse styles and rich poetic imagery. Using this dataset, we systematically evaluated decision-based and probability-based detectors, exposing significant limitations in current detection methodologies. Our results show that all decision-based detectors failed to accurately identify AI-generated poems; while certain probability-based detectors outperformed them, they still struggled significantly against our critique-based refinement strategy. Interestingly, this performance degradation was mitigated as the sample volume per set increased. Furthermore, we observed a unique phenomenon: unlike their performance on general text, LLMs exhibit a notable inability to recognize their own poetic creations. By establishing ChangAn as a rigorous benchmark, we open new avenues for research in the evolving landscape of classical Chinese poetry detection.

\section*{Limitations}
We select four representative LLMs as primary generation sources. While these models are industry standards, our findings may not generalize to all generative systems due to variations in architecture, training corpora, and fine-tuning. Additionally, our evaluation is limited to specific prominent detection tools; thus, some highly effective or emerging techniques may not have been captured in the current experiments.

\section*{Ethics Statement}
The poetry samples in this study were collected from public collections and online communities, restricted strictly to non-commercial scientific research. To ensure transparency and scholarly integrity, all human-written poems are presented with their original author names (mostly pseudonyms). By providing explicit attribution, we acknowledge the intellectual property of contemporary poets while establishing a verifiable foundation for our corpus.

Our practices strictly comply with Article 24 of the Copyright Law of the People's Republic of China through: (1) Proper Attribution, by explicitly stating authors and titles; (2) Academic Purpose, where data is used solely for research and benchmarking; and (3) Non-Commercial Nature, ensuring that scholarly sharing for reproducibility does not compete with the works' normal exploitation or prejudice the holders' legitimate interests. This open-source contribution ultimately facilitates the intersection of technology and literature. 

\section*{Acknowledgments}
This work was funded by National Natural Science Foundation of China (Grant No. 62366036), Outstanding Youth Fund Project of Inner Mongolia Autonomous Region( Grant No. 2025JQ010), Program for Young Talents of Science and Technology in Universities of Inner Mongolia Autonomous Region (Grant No. NJYT24033), Major Science and Technology Projects of Inner Mongolia Autonomous Region (Grant No. 2025ZDSF0029), Key R\&D and Achievement Transformation Program of Inner Mongolia Autonomous Region (Grant No. 2025YFDZ0011, 2025YFDZ0026, 2025YFSH0021, 2025YFHH0073), Hohhot Science and Technology Project (Grant No. 2023-Zhan-Zhong-1).

\bibliography{acl_latex}
\newpage
\appendix
\begin{CJK}{UTF8}{gbsn}
\label{sec:appendix}
\section{Examples of Poetry}
\label{app:examples-of-poetry}

To provide a qualitative overview of the generative capabilities, we present several examples of classical Chinese poetry authored by both humans and various LLMs in Figure \ref{fig:poetry-examples-comparison}. These samples illustrate a spectrum of styles. While the AI models demonstrate a sophisticated grasp of traditional imagery and formal constraints, the human examples often exhibit more specific, lived-experience-based narratives.

\begin{figure*}[ht]
\begin{tcolorbox}[colback=gray!5,
                  colframe=black,
                  width=0.95\textwidth]

\textbf{Author:} Human \\
\textbf{《鹧鸪天·乙巳六月代韩君忆旧并序（其四：在家中）》} \\
弃笔丢诗三载霜，光阴裁作柳丝长。抱猫闲数轩窗影，呵手空摹旧墨行。 \\
听更断，抚纹凉，晴来犹怯试春裳。偶然拾得残笺句，始觉心魂未死僵。 \\

\hrule
\vspace{0.6em}

\textbf{Author:} Deepseek-V3.2 \\
\textbf{《八声甘州·赋临沂》} \\
望沂河九曲绕琅琊，翠色接云平。正春山叠黛，银鸥点素，禹甸钟灵。\\
漫说羲之故里，翰墨染孤亭。兰渚流芳处，犹响琴声。\\
遥忆蒙山烽火，有崮崖铁骨，月照连营。叹兵书竹简，犹记汉秦旌。\\
问汤汤、千秋逝水，几回看、墟里起新城。凭栏处，沐风沂咏，一片澄明。\\

\hrule
\vspace{0.6em}

\textbf{Author:} Kimi-K2 \\
\textbf{《绮怀其十》} \\
旧读相期薄彻侯，旋抛翠黛代深愁。 \\
讵因买妇轻恩薄，翻使兰台托远游。 \\
秋尽玉炉销绛蜡，天寒芦雪压白头。 \\
情笺写剩空题壁，泼茗余香一夕休。 \\

\hrule
\vspace{0.6em}

\textbf{Author:} Doubao Seed-1.6 \\
\textbf{《浣溪沙·高铁上》} \\
铁马奔雷向远城，晴窗过眼万峰青，云边雁影逐车轻。\\
千里关河犹咫尺，一鞭风雪入烟汀，茶烟未冷已闻声。\\

\hrule
\vspace{0.6em}

\textbf{Author:} GPT-4.1 \\
\textbf{《七绝》} \\
碧落清辉照小楼，疏星几点夜如钩。\\
风翻翠叶寻幽径，月落寒烟逐水流。
\end{tcolorbox}
\caption{Examples of classical Chinese poetry from human and various LLMs.}
\label{fig:poetry-examples-comparison}
\end{figure*}

\section{Prompts We Used in ChangAn}

\label{app:prompts}
\subsection{Decision-based Detector}
For decision-based detectors, we utilize an expert-based prompting strategy. As illustrated in Figure \ref{fig:prompt-for-dbd}, we instruct the model to act as a poetry identification expert. The model is required to provide a brief analysis of the poem's linguistic and metrical features before assigning a binary label (0 for Human and 1 for AI). This process is designed to improve the reliability of the classification by forcing the model to justify its decision based on specific poetic nuances.

\begin{figure}[ht]
\begin{tcolorbox}[colback=gray!5,
                  colframe=black,
                  width=0.95\columnwidth]

\textbf{System Prompt:} \\
你是一个诗词鉴定专家。 \\
\textit{(Translation: You are an expert in classical poetry identification.)}

\vspace{0.8em}
\textbf{User Prompt Template:} \\
请判断这首诗是人类写的还是AI写的。请先简单分析，最后回答0（代表人类）或1（代表AI）。\\
诗词内容如下：\textcolor{blue}{\{poem\_text\}}

\vspace{0.8em}
\textbf{User Prompt Template (English):} \\
Please judge whether this poem is written by a human or an AI. Provide a brief analysis, and finally answer 0 (for human) or 1 (for AI). \\
The poem content is as follows: \textcolor{blue}{\{poem\_text\}}

\end{tcolorbox}

\caption{Prompts for generation, refinement, and detection}
\label{fig:prompt-for-dbd}
\end{figure}

\subsection{Poetry Generation}
In this section, we present the detailed configurations of the two prompting strategies employed in our study: the Direct Generation Prompt ($P_g$) and the Critique-driven Refinement Prompt ($P_c$). The $P_g$ is designed to evaluate the capability of LLMs in direct creation under strict metrical and stylistic constraints based on provided titles. 

Conversely, the $P_c$ implements an iterative refinement process where the model first identifies the given poem's shortcomings in metrics, word choice, coherence of imagery, and novelty, and subsequently performs revisions to achieve higher artistic quality. The prompts can be found in Figure \ref{fig:poetry-prompts}; both the original Chinese prompts and English translation are provided.

\begin{figure*}[ht]
\begin{tcolorbox}[colback=gray!5,
                  colframe=black,
                  width=0.95\textwidth]

\textbf{Direct Generation Prompt ($P_g$):} \\
你是一位精通中国古典诗词的专家。请你以 \textcolor{blue}{\{title\}} 为题创作一首诗词。\\
要求：严格遵守格律（平仄、押韵、对仗），严禁使用现代汉语口语。\\
输出格式：仅以 JSON 格式输出，包含 title 和 content 字段。\\

\textbf{English Translation}\\
You are an expert in classical Chinese poetry. Please create a poem titled \textcolor{blue}{\{title\}}. \\
Requirements: Strictly adhere to metrical rules including tonal patterns (Pingze), rhyme schemes, and parallelism (Duizhang). The use of modern vernacular Chinese is strictly prohibited. \\
Output Format: Provide the result only in JSON format containing the fields "title" and "content". \\

\vspace{0.8em}
\hrule
\vspace{0.8em}

\textbf{Critique-driven Refinement Prompt ($P_c$):} \\
\textbf{Round 1 (Critique):} \\
请阅读这首诗词，从格律、炼字、意境连贯性及新颖性四个维度指出其不足之处：\\
\textcolor{blue}{[标题：\{title\}] [正文：\{text\}]} \\

\textbf{Round 2 (Refinement):} \\
请根据上述建议对原诗词进行深度修改，确保作品符合格律规范。\\
输出格式：仅以 JSON 格式输出，包含 title 和 content 字段。\\

\textbf{English Translation}\\
\textbf{Round 1 (Critique):} \\
Identify the shortcomings of this poem across four dimensions, specifically metrics, word choice, coherence of imagery, and novelty: \\
\textcolor{blue}{[Title: \{title\}] [Text: \{text\}]} \\

\textbf{Round 2 (Refinement):} \\
Perform a deep revision based on the identified weaknesses to ensure the work perfectly complies with metrical standards. \\
Output Format: Provide the result only in JSON format containing the fields "title" and "content".

\end{tcolorbox}

\caption{Prompts for direct generation and critique-driven refinement}
\label{fig:poetry-prompts}

\end{figure*}

\section{A Comparative Analysis of Creative Preferences between Human and AI}
\label{app:imagerywords}

This section explains the style of using imagery and semantic differences between AI-generated and human-written poetry based on the visual and statistical evidence.

\begin{figure*}[t]
    \centering
    \begin{subfigure}[b]{0.48\textwidth}
        \centering
        \includegraphics[width=\textwidth]{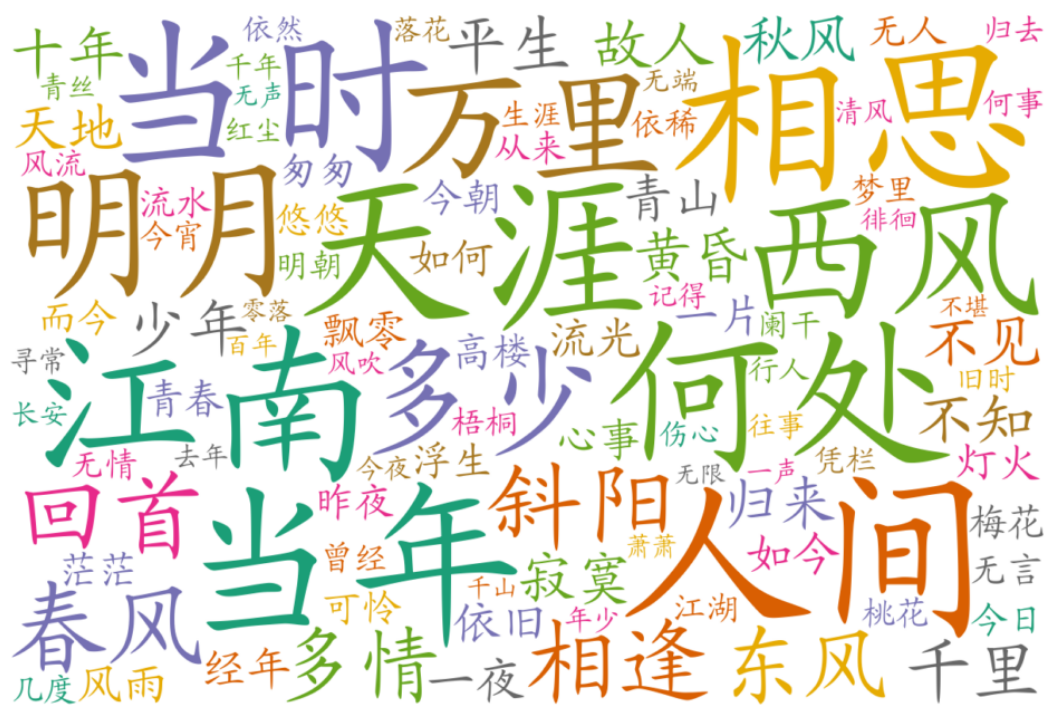}
        \caption{Human (Short Imagery Words)}
        \label{fig:h2}
    \end{subfigure}
    \hfill
    \begin{subfigure}[b]{0.48\textwidth}
        \centering
        \includegraphics[width=\textwidth]{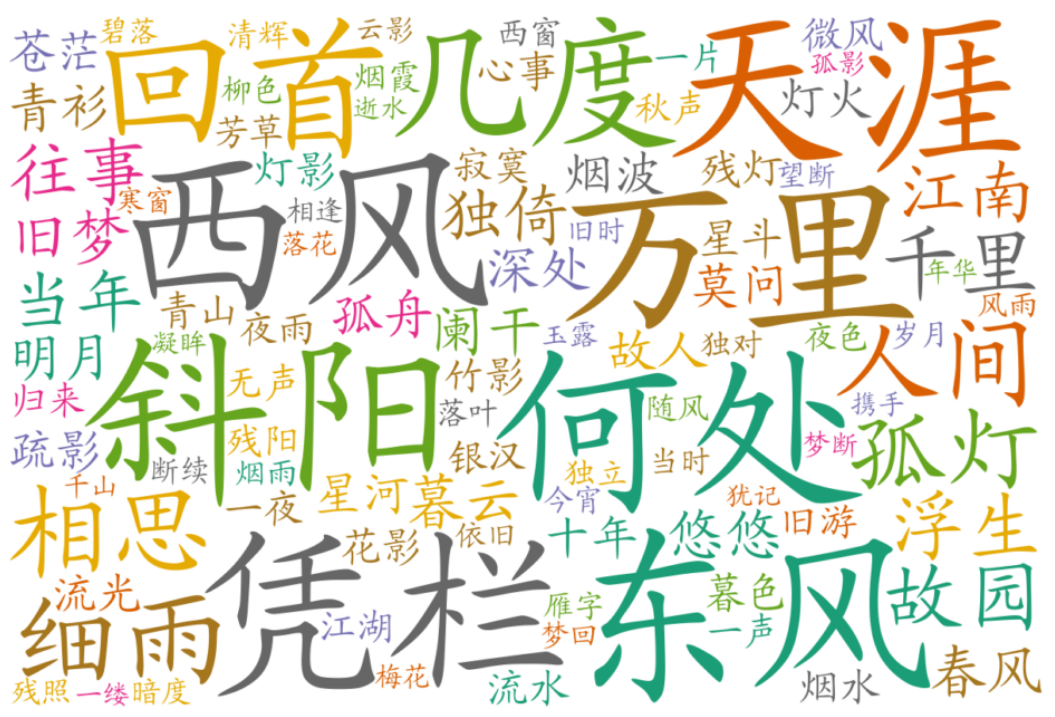}
        \caption{AI (Short Imagery Words)}
        \label{fig:ai2}
    \end{subfigure}

    \vspace{1em} 

    \begin{subfigure}[b]{0.48\textwidth}
        \centering
        \includegraphics[width=\textwidth]{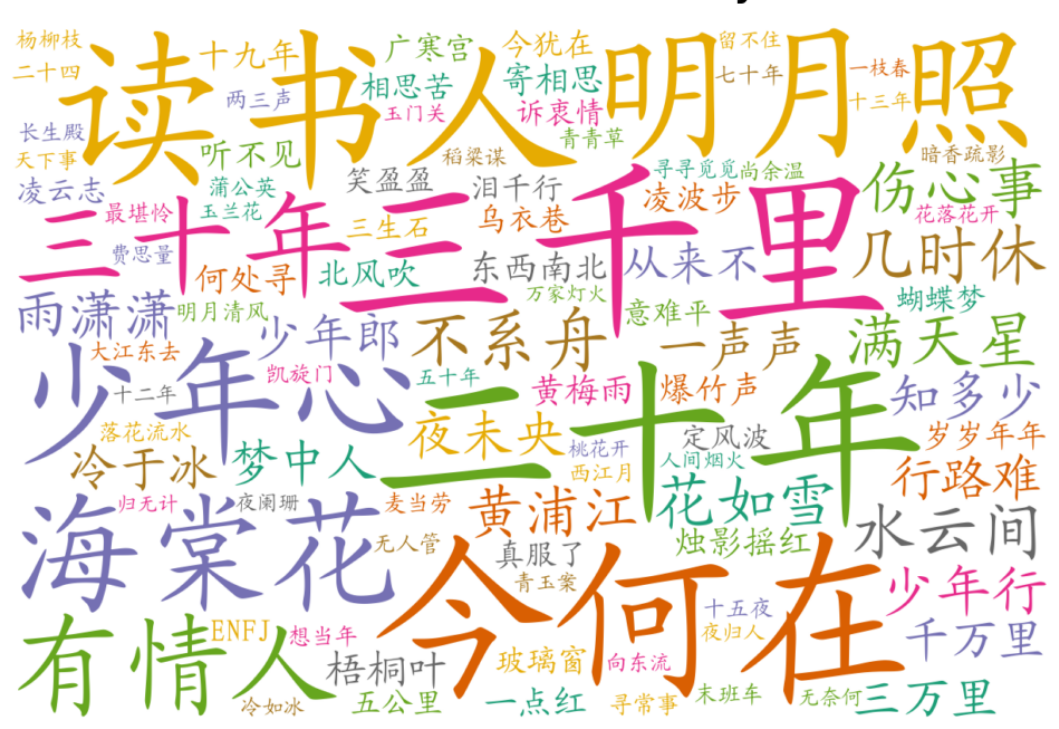}
        \caption{Human (Long Imagery Words)}
        \label{fig:h34}
    \end{subfigure}
    \hfill
    \begin{subfigure}[b]{0.48\textwidth}
        \centering
        \includegraphics[width=\textwidth]{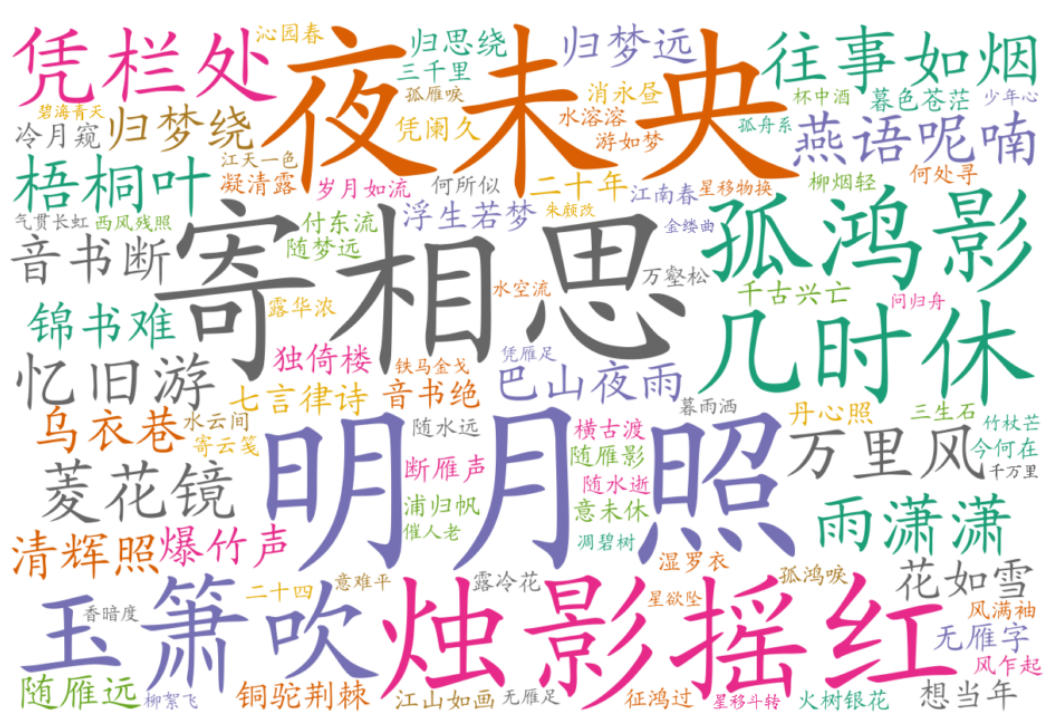}
        \caption{AI (Long Imagery Wordss)}
        \label{fig:ai34}
    \end{subfigure}

    \caption{Visual comparison of poetic imagery distributions. The left column (a, c) shows human compositions, while the right column (b, d) represents AI-generated poetry. }
    \label{fig:word-distribution-comparison}
\end{figure*}

As shown in Figure~\ref{fig:word-distribution-comparison} and Table~\ref{tab:freq-comparison-eng}, AI and human authors exhibit high similarity in their choice of common imagery. Both groups frequently utilize terms such as 西风, 相思, 江南, and 天涯. This indicates that LLMs have mastered the fundamental imagery vocabulary of classical poetry, making it difficult to distinguish authorship based solely on high-frequency imagery. 

\begin{figure*}[htbp]
    \centering
    \includegraphics[width=0.95\textwidth]{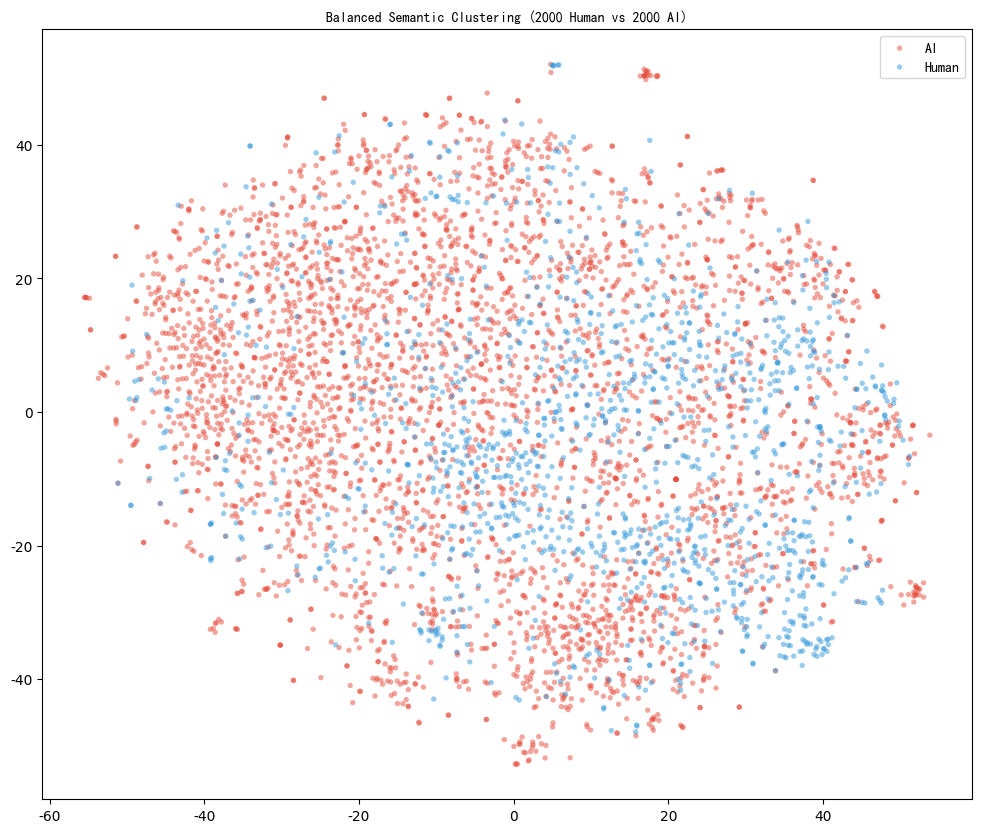}
    \caption{Semantic Clustering of AI vs. Human Poetry}
    \label{fig:cluster}
\end{figure*}

However, the semantic clustering of 2000 human-written and 2000 AI-generated poems in Figure~\ref{fig:cluster} reveals distinct distributional differences, where specific areas are occupied almost exclusively by human-written poetry.

Qualitative analysis of poems in these human-exclusive zones shows that human authors frequently include specific details, such as microscopic action descriptions, biological names (e.g., 狸花, 蜻蜓), or onomatopoeia (e.g., 滴答). AI-generated works, conversely, often remain confined to vague and generalized descriptions.

Additionally, AI employs broad, poetic-sounding terms like 万里, 凭栏, and 几度 significantly more often than human authors. Conversely, human vocabulary includes more precise spatial and temporal expressions, such as 三千里 or 二十年. These differences in lexical usage explain why human poetry exhibits greater semantic breadth than AI-generated outputs.

\begin{table*}[t]
\centering
\resizebox{\textwidth}{!}{
\setlength{\tabcolsep}{2pt} 
\begin{tabular}{rlcrlc | rlcrlc}
\toprule
\multicolumn{6}{c|}{\textbf{Top 20 High-Frequency Short Imagery Words}} & \multicolumn{6}{c}{\textbf{Top 20 High-Frequency Long Imagery Words}} \\
\cmidrule(lr){1-6} \cmidrule(lr){7-12}
\textbf{Rank} & \textbf{Human} & \textbf{Count} & \textbf{Rank} & \textbf{AI} & \textbf{Count} & \textbf{Rank} & \textbf{Human} & \textbf{Count} & \textbf{Rank} & \textbf{AI} & \textbf{Count} \\ 
\midrule
1 & 人间 (World) & 466 & 1 & 凭栏 (Leaning) & 1035 & 1 & 三千里 (3k Miles) & 22 & 1 & 明月照 (Moonlight) & 122 \\
2 & 相思 (Longing) & 371 & 2 & 万里 (10k Miles) & 1030 & 2 & 二十年 (20 Years) & 19 & 2 & 寄相思 (Send Love) & 108 \\
3 & 天涯 (Horizon) & 369 & 3 & 何处 (Where) & 1019 & 3 & 今何在 (Where now) & 17 & 3 & 夜未央 (Deep night) & 100 \\
4 & 当年 (Past days) & 311 & 4 & 西风 (W. Wind) & 1010 & 4 & 读书人 (Scholar) & 17 & 4 & 烛影摇红 (Candle) & 76 \\
5 & 何处 (Where) & 310 & 5 & 斜阳 (Setting sun) & 999  & 5 & 明月照 (Moonlight) & 14 & 5 & 玉箫吹 (Flute) & 71 \\
6 & 当时 (Back then) & 300 & 6 & 天涯 (Horizon) & 938  & 6 & 少年心 (Youth heart) & 13 & 6 & 几时休 (When end) & 51 \\
7 & 明月 (Bright moon) & 286 & 7 & 东风 (E. Wind) & 896  & 7 & 海棠花 (Begonia) & 13 & 7 & 孤鸿影 (Lone swan) & 47 \\
8 & 江南 (Jiangnan) & 281 & 8 & 回首 (Looking back) & 727  & 8 & 三十年 (30 Years) & 13 & 8 & 凭栏处 (Railing) & 47 \\
9 & 西风 (W. Wind) & 267 & 9 & 几度 (How many) & 700  & 9 & 有情人 (Lovers) & 12 & 9 & 往事如烟 (Smoke) & 47 \\
10 & 万里 (10k Miles) & 263 & 10 & 细雨 (Drizzle) & 660 & 10 & 不系舟 (Untethered) & 12 & 10 & 忆旧游 (Old trip) & 39 \\
11 & 多少 (How much) & 237 & 11 & 人间 (World) & 636 & 11 & 几时休 (When end) & 12 & 11 & 雨潇潇 (Rainy) & 38 \\
12 & 春风 (Spring wind) & 233 & 12 & 相思 (Longing) & 628 & 12 & 水云间 (Clouds) & 11 & 12 & 万里风 (Strong wind) & 38 \\
13 & 斜阳 (Setting sun) & 221 & 13 & 孤灯 (Lone lamp) & 584 & 13 & 满天星 (Starry sky) & 11 & 13 & 梧桐叶 (Wutong) & 38 \\
14 & 回首 (Looking back) & 221 & 14 & 千里 (1k Miles) & 576 & 14 & 伤心事 (Sadness) & 11 & 14 & 菱花镜 (Mirror) & 34 \\
15 & 相逢 (Meeting) & 214 & 15 & 故园 (Homeland) & 574 & 15 & 花如雪 (Snowy fl.) & 11 & 15 & 燕语呢喃 (Swallows) & 33 \\
16 & 东风 (E. Wind) & 210 & 16 & 补生 (Fleeting) & 565 & 16 & 黄浦江 (HP River) & 11 & 16 & 音书断 (Lost mail) & 31 \\
17 & 多情 (Sentimental) & 203 & 17 & 往事 (Past) & 560 & 17 & 少年行 (Youth trip) & 11 & 17 & 清辉照 (Pure light) & 31 \\
18 & 千里 (1k Miles) & 202 & 18 & 江南 (Jiangnan) & 520 & 18 & 雨潇潇 (Rainy) & 11 & 18 & 乌衣巷 (Wuyi Lane) & 29 \\
19 & 平生 (Life-long) & 198 & 19 & 当年 (Past days) & 519 & 19 & 夜未央 (Deep night) & 10 & 19 & 锦书难 (Letter) & 28 \\
20 & 少年 (Youth) & 194 & 20 & 独倚 (Alone) & 515 & 20 & 知多少 (How many) & 10 & 20 & 归梦绕 (Dream) & 28 \\
\bottomrule
\end{tabular}}
\caption{Top 20 high-frequency poetic imagery (with translations) in human-written vs. AI-generated poetry.}
\label{tab:freq-comparison-eng}
\end{table*}

\section{Models' Critique}
\label{app:models-critique}
To illustrate how the critique-based refinement works, Figure \ref{fig:critique-process} presents a case. Using a human-written poem     as the input, we demonstrate how the model identifies subtle deficiencies and performs refinement.

In Round 1, the model provides an analytical critique of the original poem 《水调歌头·夜车》. It identifies specific linguistic features, such as the use of “\charYin\charYin”, which it perceives as archaic and disruptive to the reading flow, and notes the high density of imagery in lines like “萤之甸，樱之海，梦之浔”. The critique reflects the model's internal preference for stylistic consistency and metrical rigor .

In Round 2, the model generates a Refined Poem according to its own evaluative criteria. The refinement shows a clear shift in poetic strategy: the model replaces the modern mechanical metaphors (e.g., “熔金”, “发条音”) and fragmented imagery with more conventional classical motifs (e.g., “风帆”, “野渚”). This process does not necessarily imply a subjective improvement in artistic quality; rather, it reveals the model’s tendency to normalize creative expressions toward a more traditional and “harmonious” aesthetic. 

\begin{figure*}[ht]
\begin{tcolorbox}[colback=gray!5,
                  colframe=black,
                  width=0.95\textwidth]

\textbf{Round 1: Critique} \\
\textbf{Original Poem (Human):} \\
\textbf{《水调歌头·夜车》} \\
穿山以充隧，暗轨递熔金。长车虬甲蚿足，嘘吸复\charYin\charYin。刬出仙踪绿野，楔入他乡明月，驶过梦中心。以灯为起点，被夜作行襟。\\
萤之甸，樱之海，梦之浔。人间如是车节，上下各侵寻。又向荒城谜霭，默送邻铺魅客，我岂客初临？大月钟盘挂，时作发条音。\\

\vspace{0.4em}
\textbf{Critique Results:} \\
\textbf{1. 格律：存在变通与依谱的权衡}
总体合规：作品基本遵循了《水调歌头》一体的句式与韵位框架。
可议之处：部分句子的平仄处理采取了较为宽泛或变通的方式。

\textbf{2. 炼字：个别字词可更精纯}
新奇与生涩的平衡：如“\charYin\charYin”（yín，形容蠕动貌）形容车行，用字古奥，虽求新奇，但略显隔阂，影响阅读的流畅感。“行襟”（以衣襟喻行程）的比喻也稍显曲折，需要读者停顿解意。

意象密度：“萤之甸，樱之海，梦之浔”三句意象并置，密度极高，但彼此间的逻辑或情感联结稍弱，略显堆叠。

\textbf{3. 意境连贯性：线索稍显跳跃}
空间转换：上阕从“穿山”、“暗轨”的隧道实景，突然切换到“仙踪绿野”与“梦中心”的虚境，中间缺乏必要的过渡或暗示，转折略显突兀。

主旨聚焦：下阕“人间如是车节”引发哲理思考，但随后“荒城谜霭”、“邻铺魅客”的具体场景，与结尾“大月钟盘挂，时作发条音”的宏大、抽象比喻之间，意脉的衔接可以更紧密，以形成更强的整体合力。

\textbf{4. 新颖性：融合的完成度}
概念的创新：将现代“夜车”旅程与古典词牌结合，并注入“熔金”、“钟盘发条”等工业、机械意象，构思本身具有强烈的现代性和实验精神。

融合的挑战：部分现代或奇幻意象（如“熔金”、“发条音”）与词中“仙踪”、“魅客”等古典幽玄意象，以及“萤甸”、“樱海”等自然意象，尚未完全熔铸成一个气息贯通、自洽浑成的独立意境世界，稍欠圆融。

\vspace{0.8em}
\hrule
\vspace{0.8em}

\textbf{Round 2: Refinement} \\
\textbf{Refined Poem:} \\
\textbf{《水调歌头·夜车》} \\
穿隧夜车急，金轨照霜深。铁龙流影千里，风急万山阴。灯火星辰对语，客梦风帆暂泊，远度月随心。此夜三更冷，万象寂无音。\\
晓色浅，星欲坠，路犹斟。长歌一枕南北，陌上故园寻。草木潜移故径，云雾低垂野渚，人事共沉沉。谁识轮蹄远，照我旧乡林。

\end{tcolorbox}
\caption{The example of the process of critique-driven refinement.}
\label{fig:critique-process}
\end{figure*}

\subsection{Human Performance}
\label{app:human-performance}

To establish a human baseline and assess the detection difficulty for human readers, we recruited four participants (They are not professional poets. However, they have received over ten years of Chinese classical poetry and literature education before university.) for the evaluation. Given the high cognitive load required for distinguishing classical poetry, we randomly sampled 600 identical poems (approximately 2\% of the total dataset) for each participant. While the percentage might seem modest, the absolute number of samples is sufficient to support a robust human performance. Their individual and average performances are summarized in the table below, where we report the score of Accuracy, Recall (AI/Human), Macro-Recall and Macro-F1.
\begin{table*}[t]
  \centering
  \resizebox{0.95\textwidth}{!}{ 
  \begin{tabular}{lcccccc}
    \hline
    \rowcolor{blue!15} \textbf{Human Readers} & \textbf{Acc.} & \textbf{Recall (AI)} & \textbf{Recall (Human)} & \textbf{Macro-Recall} & \textbf{Macro-F1} \\
    \hline
    
    HR1   & 47.00	&66.67&13.51	&40.09&	38.59 \\
    \rowcolor{blue!5} HR2 & 51.67	&76.19&21.05&	48.62&	39.57 \\
    HR3 & 52.50	&69.44&8.69	&39.07	&36.04 \\
    \rowcolor{blue!5} HR4 & 53.83&	74.28&16.67&	45.48&	42.91 \\

    \hline
  \end{tabular}}
  
  \caption{The Human Performance on the ChangAn Dataset.}
  \label{tab:human-performance}
\end{table*}

As shown in Table~\ref{tab:human-performance} The accuracy of the four human participants is concentrated between 47.00\% and 53.83\%, the average Accuracy is around 51.25\%, which is very close to random guessing (50\%). This indicates that even for readers with over ten years of literature education, distinguishing between AI and human classical poetry is extremely difficult.

Besides, the low Macro-F1 scores (average 39.28\%, slightly better than decision-based detectors but worse than probability-based detectors) further confirm that human judgement is unreliable for this task. These results strongly support the necessity of developing automated detection methods, as human judgment is no longer sufficient to safeguard the authenticity of classical poetry.


\end{CJK}

\end{document}